\documentclass{ecai2014}
\usepackage{times}
\usepackage{graphicx}
\usepackage{latexsym}
\usepackage{amsmath}  
\usepackage{verbatim} 
\usepackage{algorithm} 
\usepackage{algorithmic} 
\usepackage{amsfonts}
\usepackage{amssymb}
\usepackage{longtable}
\usepackage{graphicx}
\usepackage{booktabs}

\usepackage{float}
\usepackage{graphicx}
\usepackage{caption}
\usepackage{subcaption}
\usepackage{multirow}
\usepackage{url}
\newcommand{\urlBiBTeX}[1]{\url{#1}}

\floatname{algorithm}{Listing}

\DeclareMathOperator{\val}{=}  

\def\happensAt{\textsf{\footnotesize happensAt}}
\def\happens{\textsf{\footnotesize happensAt}}

\def\holdsAt{\textsf{\footnotesize holdsAt}}
\def\holdsFor{\textsf{\footnotesize holdsFor}}

\def\initiatedAt{\textsf{\footnotesize initiatedAt}}
\def\terminatedAt{\textsf{\footnotesize terminatedAt}}

\def\broken{\textsf{\footnotesize broken}}
\def\startE{\textsf{\footnotesize start}}
\def\endE{\textsf{\footnotesize end}}

\def\unionall{\textsf{\footnotesize union\_all}}

\def\intersectall{\textsf{\footnotesize intersect\_all}}

\def\complementall{\textsf{\footnotesize relative\_complement\_all}}

\def\true{\textsf{\footnotesize true}}
\def\false{\textsf{\footnotesize false}}

\def\leave{$\mathit{leaving\_object}$}
\def\move{$\mathit{moving}$}

\def\WM{\textit{WM}}

\newenvironment{mysplit}%
  {\arraycolsep 0pt \begin{array}{l}}%
  {\end{array}}

\begin{document}

\title{Reactive Reasoning with the Event Calculus}

\author{Alexander Artikis\institute{University of Piraeus, Greece \& NCSR Demokritos, Greece, email: a.artikis@unipi.gr} \and Marek Sergot\institute{Imperial College London, UK, email: m.sergot@imperial.ac.uk} \and Georgios Paliouras\institute{NCSR Demokritos, Greece, email: paliourg@iit.demokritos.gr} }

\maketitle
\bibliographystyle{ecai2014}

\begin{abstract}
Systems for symbolic event recognition accept as input a stream of time-stamped events from sensors and other computational devices, and seek to identify high-level composite events, collections of events that satisfy some pattern.
RTEC is an Event Calculus dialect with novel implementation and `windowing' techniques that allow for efficient event recognition, scalable to large data streams. 
RTEC can deal with applications where event data arrive with a (variable) delay from, and are revised by, the underlying sources. RTEC can update already recognised events and recognise new events when data arrive with a delay or following data revision.      
Our evaluation shows that RTEC can support real-time event recognition and is capable of meeting the performance requirements identified in a recent survey of event processing use cases. \footnote{A form of this paper has been submitted to IEEE TKDE.}
\end{abstract}

\section{Introduction}

Systems for symbolic event recognition (`event pattern matching') accept as input a stream of time-stamped simple, derived events (SDE)s. A SDE (`low-level event') is the result of applying a computational derivation process to some other event, such as an event coming from a sensor \cite{EPTSglossary}. Using SDEs as input, event recognition systems identify composite events (CE)s of interest---collections of events that satisfy some pattern.
The `definition' of a CE (`high-level event') imposes temporal and, possibly, atemporal constraints on its subevents, i.e.~SDEs or other CEs. Consider e.g.~the recognition of attacks on computer network nodes given the TCP/IP messages.

Numerous recognition systems have been proposed in the literature \cite{cugola11}. Recognition systems with a logic-based representation of CE definitions, in particular, have recently been attracting attention \cite{artikisKER}. They exhibit a formal, declarative semantics, in contrast to other types of recognition system that usually rely on an informal and/or procedural semantics. 
However, non-logic-based CE recognition systems have proven to be, overall, more efficient than logic-based ones. To address this issue, we present an efficient dialect of the Event Calculus \cite{kowalski86}, called `Event Calculus for Run-Time reasoning' (RTEC). The Event Calculus is a logic programming formalism for representing and reasoning about events and their effects. 
RTEC includes novel implementation techniques for efficient CE recognition, scalable to large SDE and CE volumes. 
A set of interval manipulation constructs simplify CE definitions and improve reasoning efficiency. A simple indexing mechanism makes RTEC robust to SDEs that are irrelevant to the CEs we want to recognise and so RTEC can operate without SDE filtering modules. Finally, a `windowing' mechanism supports real-time CE recognition. One main motivation for RTEC is that it should remain efficient and scalable in applications where SDEs arrive with a (variable) delay from, or are revised by, the underlying SDE detection system: RTEC can update the already recognised CEs, and recognise new CEs, when SDEs arrive with a delay or following revision. The code of RTEC is available at \url{<http://users.iit.demokritos.gr/~a.artikis/EC.html>}. 

We evaluate RTEC on public space surveillance from video content. In this application, the SDEs are the `short-term activities' detected on video frames---e.g.~a person walking, running or being inactive. The aim then is to recognise `long-term activities', i.e.~short-term activity combinations, such as when a person leaves an object unattended, when two people are moving together, when they are having a meeting or fighting. 
The CE definitions are quite complex, allowing for a realistic  evaluation of the efficiency of RTEC. This is in contrast to the majority of related approaches where rather simple CE definitions are used for empirical analysis. Our evaluation shows that RTEC supports real-time CE recognition and is capable of meeting the performance requirements of most of today's applications as estimated by a recent survey of event processing use cases~\cite{bizzaro11}.

The remainder of the paper is structured as follows. Sections \ref{sec:ec} and \ref{sec:caching} present the expressivity of RTEC and the way it performs reasoning. The experimental evaluation is given in Section \ref{sec:experiments}. Section \ref{sec:summary} summarises the presented work, puts the work in context, and outlines directions for further research.

\section{Event Calculus}\label{sec:ec}
Our system for CE recognition is based on an Event Calculus dialect. The Event Calculus \cite{kowalski86} is a logic programming formalism for representing and reasoning about events and their effects. For the dialect introduced here, called RTEC, the time model is linear and includes integer time-points. Variables start with an upper-case letter, while predicates and constants start with a lower-case letter. Where $F$ is a \emph{fluent}---a property that is allowed to have different values at different points in time---the term $F \val V$ denotes that fluent $F$ has value $V$. Boolean fluents are a special case in which the possible values are \true\ and \false. $\holdsAt(F \val V, T)$ represents that fluent $F$ has value $V$ at a particular time-point $T$. $\holdsFor(F \val V, I)$ represents that $I$ is the list of the maximal intervals for which $F\val V$ holds continuously. \holdsAt\ and \holdsFor\ are defined in such a way that, for any fluent $F$, \holdsAt$(F \val V, T)$ if and only if $T$ belongs 
to one of the maximal intervals of $I$ for which \holdsFor$(F \val V, I)$. 

An \emph{event description} in RTEC includes rules that define the event instances with the use of the \happensAt\ predicate, the effects of events with the use of the \initiatedAt\ and \terminatedAt\ predicates, and the values of the fluents with the use of the \holdsAt\ and \holdsFor\ predicates, as well as other, possibly atemporal, constraints. Table~\ref{tbl:ec} summarises the RTEC predicates available to the event description developer. The last three items in the table are interval manipulation predicates specific to RTEC. 

\begin{table}
\caption{Main predicates of RTEC.}\label{tbl:ec}
\begin{center}
\renewcommand{\arraystretch}{0.9}
\setlength\tabcolsep{3.6pt}
\begin{tabular}{ll}
\hline\noalign{\smallskip}
\multicolumn{1}{c}{\textbf{Predicate}} & \multicolumn{1}{c}{\textbf{Meaning}}  \\
\noalign{\smallskip}
\hline
\noalign{\smallskip}
\textsf{\footnotesize happensAt}$(E, T)$ & Event $E$ occurs at time $T$  \\[3pt]


\textsf{\footnotesize holdsAt}$(F \val V, T)$ & The value of fluent $F$ is $V$ at time $T$ \\[3pt]

\textsf{\footnotesize holdsFor}$(F \val V, I)$ & $I$ is the list of the maximal intervals \\
                           & for which $F\val V$ holds continuously\\[3pt]

\textsf{\footnotesize initiatedAt}$(F \val V, T)$ & At time $T$ a period of time for which\\
& $F\val V$ is initiated \\[3pt]

\textsf{\footnotesize terminatedAt}$(F \val V, T)$ & At time $T$ a period of time for which \\
& $F\val V$ is terminated \\[3pt]

\textsf{\footnotesize relative\_} 		& $I$ is the list of maximal intervals produced \\
\textsf{complement\_}	& by the relative complement of the list   \\
\textsf{all\_}$\mathit{(I', L, I)}$				& of maximal intervals $I'$ with respect to \\
 & every list of maximal intervals of list $L$ \\[3pt]

\textsf{\footnotesize union\_all}$\mathit{(L, I)}$ & $\mathit{I}$ is the list of maximal intervals  \\
			    & produced by the union of the lists of \\
& maximal intervals of list $L$ \\[3pt]

\textsf{\footnotesize intersect\_all}$\mathit{(L, I)}$ & $\mathit{I}$ is the list of maximal intervals \\
				& produced by the intersection of \\
& the lists of maximal intervals of list $\mathit{L}$ \\

\hline
\end{tabular}
\end{center}
\end{table}

We represent instantaneous SDEs and CEs by means of \happensAt, while durative SDEs and CEs are represented as fluents. The majority of CEs are durative and, therefore, in CE recognition the task generally is to compute the maximal intervals for which a fluent representing a CE has a particular value continuously. 

\subsection{Simple Fluents}\label{sec:simple-fluents}

Fluents in RTEC are \emph{simple} or \emph{statically determined}. We assume, without loss of generality, that these types are disjoint. For a simple fluent $F$, $F\val V$ holds at a particular time-point $T$ if $F \val V$ has been \emph{initiated} by an event that has occurred at some time-point earlier than $T$, and has not been \emph{terminated} at some other time-point in the meantime. This is an implementation of the law of inertia. 
To compute the \emph{intervals} $I$ for which $F\val V$, i.e.~$\holdsFor(F \val V, I)$, we find all time-points $T_s$ at which $F\val V$ is initiated, and then, for each $T_s$, we compute the first time-point $T_f$ after $T_s$ at which $F\val V$ is `broken'.
The time-points at which $F\val V$ is initiated are computed by means of domain-specific \initiatedAt\ rules.
The time-points at which $F\val V$ is `broken' are computed as follows:
\begin{align}
& \label{eq:ec-broken-terminated}
\begin{mysplit}
\broken(F\val V,\ T_s,\ T) \leftarrow \\
\quad   \terminatedAt(F\val V,\ T_f),\ \ T_s < T_f \leq T
\end{mysplit}\\
& \label{eq:ec-broken-initiated}
\begin{mysplit}
\broken( F\val V_1,\ T_s,\ T) \leftarrow \\
\quad  \initiatedAt(F\val V_2,\ T_f),\ \ T_s < T_f \leq T,\ \ V_1 \neq V_2    
\end{mysplit}
\end{align}
$\broken(F\val V, T_s, T)$ represents that a maximal interval starting at $T_s$ for which
$F\val V$ holds continuously is terminated at some time $T_f$ such that $T_s{<}T_f{\leq}T$.
Similar to \initiatedAt, \terminatedAt\ rules are domain-specific (examples are presented below).
According to rule \eqref{eq:ec-broken-initiated}, if $F \val V_2$ is initiated at $T_f$ then effectively $F \val V_1$ is terminated at time $T_f$, for all other possible values $V_1$ of $F$. Rule \eqref{eq:ec-broken-initiated} ensures therefore that a fluent cannot have more than one value at any time.
We do not insist that a fluent must have a value at every time-point. There is a difference between initiating a Boolean fluent $F \val \false$ and terminating $F \val \true$: the former implies, but is not implied by, the latter.

RTEC stores and indexes \holdsFor\ intervals as they are computed for any given fluent-value $F \val V$: thereafter intervals for $F \val V$ are retrieved from the computer memory without the need for re-computation. Similarly, a \holdsAt\ query for $F \val V$ looks up  $F$'s value in the \holdsFor\ cache.

In public space surveillance, it is often required to detect when a person leaves an object unattended. Typically, an object carried by a person is not tracked by the computer vision algorithms---only the person that carries it is tracked. The object will be tracked, i.e.~it will `appear', if and only if the person leaves it somewhere. Moreover, objects (as opposed to persons) can exhibit only inactive activity. Accordingly, we define a durative `leaving an object' CE as follows:
\begin{align}
& \label{eq:leaving-inactive-init}
\begin{mysplit}
\initiatedAt( \mathit{leaving\_object(P, Obj)\val\true,\ T} ) \leftarrow \\
\qquad\happens( \mathit{appear(Obj),\ T} ), \\
\qquad\holdsAt( \mathit{inactive(Obj)\val\true,\ T} ), \\
\qquad\holdsAt( \mathit{close(P, Obj)\val\true,\ T} ), \\
\qquad\holdsAt( \mathit{person(P)\val\true,\ T} )
\end{mysplit}\\
& \label{eq:leaving-exit-term}
\begin{mysplit}
\initiatedAt( \mathit{leaving\_object(P, Obj)\val\false,\ T} ) \leftarrow \\
\qquad\happens( \mathit{disappear(Obj),\ T} )
\end{mysplit}
\end{align}
In rule \eqref{eq:leaving-inactive-init} \leave$\mathit{(P, Obj)}\val\true$ is initiated at time $T$ if $\mathit{Obj}$ `appears' at $T$, it is inactive at $T$, and there is a person $P$ `close' to $\mathit{Obj}$ at $T$.  $\mathit{appear}$ and $\mathit{inactive}$ are instantaneous SDE and durative SDE respectively. SDE are detected on video frames in this application. $\mathit{close(A, B)}$ is \true\ when the distance between $A$ and $B$ does not exceed some threshold of pixel positions.

There is no explicit information about whether a tracked entity is a person or an inanimate object. 
We define the simple fluent $\mathit{person(P)}$ to have value \true\ if $P$ has been active, walking, running or moving abruptly since $P$ first `appeared'. The value of $\mathit{person(P)}$ has to be time-dependent because the identifier $P$ of a tracked entity that `disappears'  (is no longer tracked) at some point  may be used later to refer to another entity that `appears' (becomes tracked), and that other entity may not necessarily be a person. This is a feature of the application and not something that is imposed by RTEC.

Unlike the specification of $\mathit{person}$, it is not clear from the data whether a tracked entity is an object. $\mathit{person(P)\val\false}$ does not necessarily mean that $P$ is an object; it may be that $P$ is not tracked, or that $P$ is a person that has never walked, run, been active or moved abruptly. Note finally that rule~\eqref{eq:leaving-inactive-init} incorporates a reasonable simplifying assumption,  that a person entity will never exhibit `inactive' activity at the moment it first `appears' (is tracked). If an entity is `inactive' at the moment it  `appears' it can be assumed to be an object, as in the first two conditions of rule~\eqref{eq:leaving-inactive-init}.

Rule \eqref{eq:leaving-exit-term} expresses the conditions in which \leave\ ceases to be recognised. \leave$\mathit{(P,Obj)}$ becomes \false\ when the object in question is picked up.
An object that is picked up by someone is no longer tracked---it `disappears'---terminating \leave. ($\mathit{disappear}$ is an instantaneous SDE.)
The maximal intervals during which  \leave$\mathit{(P,Obj)\val\true}$ holds continuously are computed using the built-in RTEC predicate \holdsFor\ from rules \eqref{eq:leaving-inactive-init} and \eqref{eq:leaving-exit-term}.

Consider another example from public space surveillance:
\begin{align}
& \label{eq:move-simple-fluent1}
\begin{mysplit}
\initiatedAt\mathit{(moving(P_1,P_2)\val\true,\ T)} \ \leftarrow\\
\qquad  \happens\mathit{(\startE(walking(P_1)\val\true),\, T),} \\
\qquad  \holdsAt\mathit{(walking(P_2)\val\true,\, T),} \\
\qquad  \holdsAt(\mathit{close(P_1, P_2)\val\true,\ T} ) \\
\end{mysplit} 
\end{align}
\begin{align}
& \label{eq:move-simple-fluent2}
\begin{mysplit}
\initiatedAt\mathit{(moving(P_1,P_2)\val\true,\ T)} \ \leftarrow\\
\qquad  \happens\mathit{(\startE(walking(P_2)\val\true),\, T),} \\
\qquad  \holdsAt\mathit{(walking(P_1)\val\true,\, T),} \\
\qquad  \holdsAt(\mathit{close(P_1, P_2)\val\true,\ T } ) \\
\end{mysplit} \\ 
& \label{eq:move-simple-fluent3}
\begin{mysplit}
\initiatedAt\mathit{(moving(P_1,P_2)\val\true,\ T)} \ \leftarrow\\
\qquad  \happens\mathit{(\startE(close(P_1, P_2)\val\true),\, T),} \\
\qquad  \holdsAt\mathit{(walking(P_1)\val\true,\, T),} \\
\qquad  \holdsAt\mathit{(walking(P_2)\val\true,\, T)} \\
\end{mysplit}\\
& \label{eq:move-simple-fluent4}
\begin{mysplit}
\terminatedAt\mathit{(moving(P_1,P_2)\val\true,\ T)}\ \leftarrow \\
\qquad  \happens\mathit{(\endE(walking(P_1)\val\true),\, T)}\\
\end{mysplit} \\
& \label{eq:move-simple-fluent5}
\begin{mysplit}
\terminatedAt\mathit{(moving(P_1,P_2)\val\true,\ T)}\ \leftarrow \\
\qquad  \happens\mathit{(\endE(walking(P_2)\val\true),\, T)}\\
\end{mysplit} \\
& \label{eq:move-simple-fluent6}
\begin{mysplit}
\terminatedAt\mathit{(moving(P_1,P_2)\val\true,\ T)}\ \leftarrow \\
\qquad  \happens\mathit{(\endE(close(P_1, P_2)\val\true),\, T)}
\end{mysplit}
\end{align}
$\mathit{walking}$ is a durative SDE detected on video frames. 
$\startE(F\val V)$ (resp.~$\endE(F\val V)$) is a built-in RTEC event taking place at each starting (ending) point of each maximal interval for which $F\val V$ holds continuously.
The above formalisation states that $P_1$ is moving with $P_2$ when they are walking close to each other.

One of the main attractions of RTEC is that it makes available the power of logic programming to express complex temporal and atemporal constraints, as conditions in \initiatedAt\ and \terminatedAt\ rules for durative CEs, and \happensAt\ rules for instantaneous CEs. E.g.~standard event algebra operators, such as sequence, disjunction, parallelism, etc, may be expressed in a RTEC event description.

\subsection{Statically Determined Fluents}\label{sec:sd-fluents}

In addition to the domain-independent definition of \holdsFor, an event description may include domain-specific \holdsFor\ rules, used to define the values of a fluent $F$ in terms of the values of other fluents. We call such a fluent $F$ \emph{statically determined}. 
\holdsFor\ rules of this kind make use of interval manipulation constructs---see the last three items of Table \ref{tbl:ec}. Consider, e.g.~\move\ as in rules~\eqref{eq:move-simple-fluent1}--\eqref{eq:move-simple-fluent6} but defined instead as a statically determined fluent:
\begin{align}
& \label{eq:move}
\begin{mysplit}
\holdsFor( \mathit{moving(P_1, P_2)\val\true,\ I } ) \leftarrow \\
\qquad \mathit{\holdsFor( walking(P_1)\val\true,\ I_1 ),} \\
\qquad \mathit{\holdsFor( walking(P_2)\val\true,\ I_2 ),} \\
\qquad\holdsFor( \mathit{close(P_1, P_2)\val\true,\ I_3 } ), \\
\qquad \intersectall\mathit{( [I_1, I_2, I_3],\ I )}
\end{mysplit}
\end{align}
The list $I$ of maximal intervals during which $P_1$ is moving with $P_2$ is computed by determining the list $I_1$ of maximal intervals during which $P_1$ is walking, the list $I_2$ of maximal intervals during which $P_2$ is walking, the list $I_3$ of maximal intervals during which $P_1$ is close to $P_2$, and then calculating the list $I$ representing the intersections of the maximal intervals in $I_1$, $I_2$ and $I_3$. 

RTEC provides three interval manipulation constructs: \unionall, \intersectall\ and \complementall. $\unionall(L, I)$ computes the list $I$ of maximal intervals representing the union of maximal intervals of the lists of list $L$. For instance: 
\begin{align}
& \nonumber
\begin{mysplit}
\unionall([[(5,20), (26,30)],[(28,35)]],\ [(5,20), (26,35)])
\end{mysplit}
\end{align}
A term of the form $\mathit{(T_s, T_e)}$ in RTEC represents the closed-open interval $\mathit{[T_s, T_e)}$. $I$ in \unionall$(L, I)$ is a list of maximal intervals that includes each time-point that is part of at least one list of $L$. 

$\intersectall(L, I)$ computes the list $I$ of maximal intervals such that $I$ represents the intersection of maximal intervals of the lists of list $L$, as, e.g.:
\begin{align}
& \nonumber
\begin{mysplit}
\intersectall([[(26,31)], [(21,26),(30,40)]],\ [(30,31)])
\end{mysplit}
\end{align}
$I$ in \intersectall$(L, I)$ is a list of maximal intervals that includes each time-point that is part of all lists of $L$.

$\complementall(I', L, I)$ computes the list $I$ of maximal intervals such that $I$ represents the relative complements of the list of maximal intervals $I'$ with respect to the maximal intervals of the lists of list $L$. 
Below is an example of \complementall:
\begin{align}
& \nonumber
\begin{mysplit}
\complementall([(5,20), (26,50)],\\
\quad [[(1,4),(18,22)],[(28,35)]],\ [(5,18),(26,28),(35,50)])
\end{mysplit}
\end{align}
$I$ in \complementall$(I', L, I)$ is a list of maximal intervals that includes each time-point of $I'$ that is not part of any list of $L$.

When defining a statically determined fluent $F$ we will often want to say that, for all time-points $T$, $F \val V$ holds at $T$ if and only if $W$ holds at $T$ where $W$ is some Boolean combination of fluent-value pairs. RTEC provides optional shorthands for writing such definitions concisely. For example, the definition
\begin{align}
& \label{eq:iff-example}
\begin{mysplit}
G \val V\ \mathsf{iff} \\
\qquad (A \val V_1 \ \mathsf{or} \ B \val V_2), \\
\qquad (A \val V_1' \ \mathsf{or} \ B \val V_2'), \\
\qquad \mathsf{not}\ C \val V_3
\end{mysplit}
\end{align}
is expanded into the following \holdsFor\ rule:
\begin{align}
& \label{eq:iff-example-full}
\begin{mysplit}
\holdsFor(G\val V,\ I)\ \leftarrow \\
\qquad  \holdsFor(A\val V_1,\ I_1), \
        \holdsFor(B\val V_2,\ I_2), \\
\qquad  \unionall([I_1, I_2],\ I_3), \\
\qquad  \holdsFor(A\val V_1',\ I_4), \
        \holdsFor(B\val V_2',\ I_5), \\
\qquad  \unionall([I_4, I_5],\ I_6), \\
\qquad  \intersectall([I_3,I_6],\ I_7), \\
\qquad  \holdsFor(C\val V_3,\ I_8), \\
\qquad  \complementall(I_7,\ [I_8],\ I)
\end{mysplit}
\end{align}
The required transformation takes place automatically when event descriptions are loaded into RTEC.

For a wide range of fluents, the use of interval manipulation constructs leads to a much more concise definition than the traditional style of Event Calculus representation, i.e.~identifying the various conditions under which the fluent is initiated and terminated so that maximal intervals can then be computed using the domain-independent \holdsFor. Compare, e.g.~the statically determined and simple fluent representations of \move\ in rules~\eqref{eq:move} and \eqref{eq:move-simple-fluent1}--\eqref{eq:move-simple-fluent6} respectively.

The interval manipulation constructs of RTEC can also lead to much more efficient computation. The complexity analysis may be found in \cite{artikisDEBS12}.

\subsection{Semantics}

CE definitions are (locally) stratified logic programs \cite{local-strat}.
We restrict attention to \emph{hierarchical} definitions, those where it is possible to define a function \emph{level} that maps all fluent-values $F\val V$ and all events to the non-negative integers as follows. 
Events and statically determined fluent-values $F \val V$  of level $0$ are those whose \happensAt\ and \holdsFor\ definitions do not depend on any other events or fluents. In CE recognition, they represent the input SDEs. There are no fluent-values $F \val V$ of simple fluents $F$ in level $0$. 
Events and simple fluent-values of level $n$ are defined in terms of at least one event or fluent-value of level $n{-}1$ and a possibly empty set of events and fluent-values from levels lower than $n{-}1$. Statically determined fluent-values of level $n$ are defined in terms of at least one fluent-value of level $n{-}1$ and a possibly empty set of fluent-values from levels lower than $n{-}1$. Note that fluent-values $F\val V_i$ and $F\val V_j$ for $V_i{\neq}V_j$ could be mapped to different levels. For simplicity however, and without loss of generality, a fluent $F$ itself is either simple or statically determined but not both. The CE definitions of public space surveillance, i.e.~the \holdsFor\ definitions of statically determined fluents, \initiatedAt\ and \terminatedAt\ definitions of simple fluents and \happensAt\ definitions of events, are available with the RTEC code.

\section{Run-Time Recognition}\label{sec:caching}

CE recognition has to be efficient enough to support real-time decision-making, and scale to very large numbers of SDEs and CEs. SDEs may not necessarily arrive at the CE recognition system in a timely manner, i.e.~there may be a (variable) delay between the time at which SDEs take place and the time at which they arrive at the CE recognition system. Moreover, SDEs may be revised, or even completely discarded in the future, as in the case where the parameters of a SDE were originally computed erroneously and are subsequently revised, or in the case of retraction of a SDE that was reported by mistake, and the mistake was realised later \cite{anicic12}. Note that SDE revision is not performed by the CE recognition system, but by the underlying SDE detection system.

RTEC performs CE recognition by computing and storing the maximal intervals of fluents and the time-points in which events occur. CE recognition takes place at specified query times $Q_1, Q_{2}, \dots$. At each $Q_i$ the SDEs that fall within a specified interval---the `working memory' ($\WM$) or `window'---are taken into consideration. All SDEs that took place before or at $Q_i{-}\WM$ are discarded. This is to make the cost of CE recognition dependent only on the $\WM$ size and not on the complete SDE history. 
The $\WM$ size, and the temporal distance between two consecutive query times --- the `step' ($Q_{i}{-}Q_{i-1}$) --- are set by the user. 

At $Q_i$, the maximal intervals computed by RTEC are those that can be derived from SDEs that occurred in the interval $(Q_{i}{-}\WM, Q_{i}]$, as recorded at time $Q_i$. When $\WM$ is longer than the inter-query step, i.e., when $Q_{i}{-}\WM {<} Q_{i-1} {<} Q_{i}$, it is possible that an SDE occurs in the interval $(Q_{i}{-}\WM, Q_{i-1}]$ but arrives at RTEC only after $Q_{i-1}$; its effects are taken into account at query time $Q_i$. And similarly for SDEs that took place in $(Q_{i}{-}\WM, Q_{i-1}]$ and were subsequently revised after $Q_{i-1}$. In the common case that SDEs arrive at RTEC with delays, or there is SDE revision,  it is preferable therefore to make $\WM$ longer than the inter-query step. 
Note that information may still be lost. Any SDEs arriving or revised between $Q_{i-1}$ and $Q_i$ are discarded at $Q_i$ if they took place before or at $Q_i{-}\WM$. 
To reduce the possibility of losing information, one may increase the $\WM$ size.
%
Doing so, however, decreases recognition efficiency. 

\begin{figure}[t]
  \begin{center}
	\includegraphics[width=.5\textwidth]{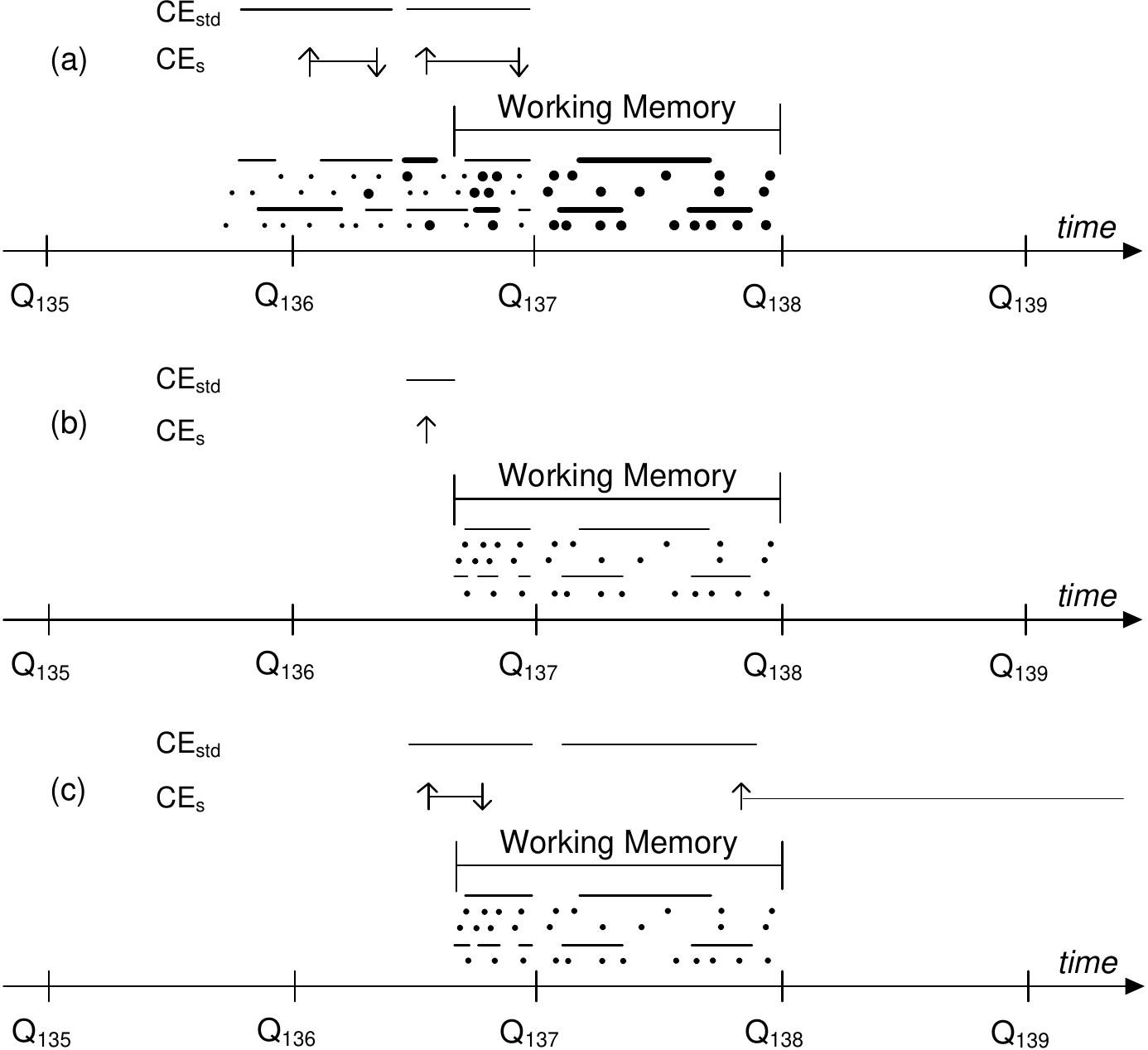}
	\caption{Windowing in RTEC.\vspace{9mm}}
	\label{fig:caching}
	\end{center}
\end{figure}

Figure \ref{fig:caching} illustrates windowing in RTEC. In this example we have  $\WM{>}Q_i{-}Q_{i-1}$. To avoid clutter, Figure \ref{fig:caching} shows streams of only five SDEs. These are displayed below $\WM$, with dots for instantaneous SDEs and lines for durative ones. For the sake of the example, we are interested in recognising just two CEs:
\begin{itemize}
 \item $\mathit{CE_{s}}$, represented as a simple fluent (see Section \ref{sec:simple-fluents}). The starting and ending points, and the maximal intervals of $\mathit{CE_{s}}$ are displayed above $\WM$ in Figure \ref{fig:caching}.
 
 \item $\mathit{CE_{std}}$, represented as a statically determined fluent (see Section \ref{sec:sd-fluents}). For the example, the maximal intervals of $\mathit{CE_{std}}$ are defined to be the union of the maximal intervals of the two durative SDEs in Figure \ref{fig:caching}.
 The maximal intervals of $\mathit{CE_{std}}$ are displayed above the $\mathit{CE_{s}}$ intervals.
\end{itemize}
For simplicity, we assume that both $\mathit{CE_{s}}$ and $\mathit{CE_{std}}$ are defined only in terms of SDE, i.e.~they are not defined in terms of other CEs.

Figure \ref{fig:caching} shows the steps that are followed in order to recognise CEs at an arbitrary query time, say $Q_{138}$. Figure \ref{fig:caching}(a) shows the state of RTEC as computation begins at $Q_{138}$. 
All SDEs that took place before or at $Q_{137}{-}\WM$ were retracted at $Q_{137}$. The thick lines and dots represent the SDEs that arrived at RTEC between $Q_{137}$ and $Q_{138}$; some of them took place before $Q_{137}$. Figure \ref{fig:caching}(a) also shows the maximal intervals for the CE fluents $\mathit{CE_{s}}$ and $\mathit{CE_{std}}$ that were computed and stored at $Q_{137}$.

The CE recognition process at $Q_{138}$ considers the SDEs that took place in $(Q_{138}{-}\WM, Q_{138}]$.
All SDEs that took place before or at $Q_{138}{-}\WM$ are discarded, as shown in Figure \ref{fig:caching}(b). For durative SDEs that started before $Q_{138}{-}\WM$ and ended after that time, RTEC retracts the sub-interval up to and including $Q_{138}{-}\WM$. Figure \ref{fig:caching}(b) shows the interval of a SDE that is partially retracted in this way. 

Now consider CE intervals. At $Q_i$ some of the maximal intervals computed at $Q_{i-1}$ might have become invalid. This is because some SDEs occurring in $(Q_i{-}\WM, Q_{i-1}]$ might have arrived or been revised after $Q_{i-1}$: their existence could not have been known at $Q_{i-1}$. Determining which CE intervals should be (partly) retracted in these circumstances can be computationally very expensive. See Section \ref{sec:summary} for a discussion. We find it simpler, and more efficient, to discard all CE intervals in $(Q_i{-}\WM, Q_i]$ and compute all intervals from scratch in that period. CE intervals that have ended before or at $Q_i{-}\WM$ are discarded. 
Depending on the user requirements, these intervals may be stored in a database for retrospective
inspection of the activities of a system.

In Figure~\ref{fig:caching}(b), the earlier of the two maximal intervals computed for $\mathit{CE_{std}}$ at $Q_{137}$ is discarded at $Q_{138}$ since its endpoint is before  $Q_{138}{-}\WM$. The later of the two intervals overlaps $Q_{138}{-}\WM$ (an interval `overlaps' a time-point $t$ if the interval starts before or at $t$ and ends after or at that time) and is partly retracted at $Q_{138}$. Its starting point  could not have been affected by SDEs arriving between $Q_{138}{-}\WM$ and $Q_{138}$ but its endpoint has to be recalculated. Accordingly, the sub-interval from $Q_{138}{-}\WM$ is retracted at $Q_{138}$.

In this example, the maximal intervals of $\mathit{CE_{std}}$ are determined by computing the union of the maximal intervals of the two durative SDEs shown in Figure \ref{fig:caching}. 
At $Q_{138}$, only the SDE intervals in $(Q_{138}{-}\WM, Q_{138}]$ are considered. 
%
In the example, there are two maximal intervals for $\mathit{CE_{std}}$ in this period as can be seen in Figure~\ref{fig:caching}(c). The earlier of them has its startpoint at $Q_{138}{-}\WM$. Since that abuts the existing, partially retracted sub-interval for $\mathit{CE_{std}}$ whose endpoint is $Q_{138}{-}\WM$, those two intervals are amalgamated into one continuous maximal interval as shown in Figure~\ref{fig:caching}(c). In this way, the endpoint of the $\mathit{CE_{std}}$ interval that overlapped $Q_{138}{-}\WM$ at $Q_{137}$ is recomputed to take account of SDEs available at $Q_{138}$. (In this particular example, it happens that the endpoint of this interval is the same as that computed at $Q_{137}$. That is merely a feature of this particular example. Had $\mathit{CE_{std}}$ been defined e.g.~as the \emph{intersection} of the maximal intervals of the two durative SDE, then the intervals of $\mathit{CE_{std}}$ would have changed in $(Q_{138}{-}\WM, Q_{137}]$.)

Figure \ref{fig:caching} also shows how the intervals of the simple fluent $\mathit{CE_{s}}$ are computed at $Q_{138}$. Arrows facing upwards (downwards) denote the starting (ending) points of $\mathit{CE_{s}}$ intervals. 
First, in analogy with the treatment of statically determined fluents, the earlier of the two $\mathit{CE_{s}}$   intervals in Figure~\ref{fig:caching}(a), and its start and endpoints, are retracted. They occur before $Q_{138}{-}\WM$. The later of the two intervals overlaps $Q_{138}{-}\WM$. The interval is retracted, and only its starting point is kept; its new endpoint, if any, will be recomputed at $Q_{138}$. See Figure \ref{fig:caching}(b). For simple fluents, it is simpler, and more efficient, to retract such intervals completely and reconstruct them later from their start and endpoints by means of the domain-independent \holdsFor\ rules, rather than keeping the sub-interval that takes place before $Q_{138}{-}\WM$, and possibly amalgamating it later with another interval, as we do for statically determined fluents.
%
%

The second step for $\mathit{CE_{s}}$ at $Q_{138}$ is to calculate its starting and ending points by evaluating the relevant \initiatedAt\ and \terminatedAt\ rules. For this, we only consider SDEs that took place in $(Q_{138}{-}\WM, Q_{138}]$. 
Figure \ref{fig:caching}(c) shows the starting and ending points of $\mathit{CE_{s}}$ in $(Q_{138}{-}\WM, Q_{138}]$. The last ending point of $\mathit{CE_{s}}$ that was computed at $Q_{137}$ was invalidated in the light of the new SDEs that became available at $Q_{138}$ (compare Figures \ref{fig:caching}(c)--(a)). Moreover, another ending point was computed at an earlier time.  

Finally, in order to recognise $\mathit{CE_{s}}$ at $Q_{138}$ we use the domain-independent \holdsFor\ to calculate the maximal intervals of $\mathit{CE_{s}}$ given its starting and ending points. The later of the  two $\mathit{CE_{s}}$ intervals computed at $Q_{137}$ became shorter when re-computed at $Q_{138}$. The second interval of $\mathit{CE_{s}}$ at $Q_{138}$ is open: given the SDEs available at $Q_{138}$, we say that $\mathit{CE_{s}}$ holds \emph{since} time $t$, where $t$ is the last starting point of $\mathit{CE_{s}}$.


The discussion above showed that, when SDEs arrive with a variable delay, CE intervals computed at an earlier query time may be (partly) retracted at the current or a future query time. (And similarly if SDEs are revised.)
Depending on the application requirements, RTEC may be set to report:

\begin{itemize}
 \item CEs as soon as they are recognised, even if their intervals may be (partly) retracted in the future.
 \item CEs whose intervals may be partly, but not completely, retracted in the future, i.e.~CEs whose intervals overlap $Q_{i+1}{-}\WM$. 
 \item CEs whose intervals will not be even partly retracted in the future, i.e.~CEs whose intervals end before or at $Q_{i+1}{-}\WM$.
\end{itemize}

The example used for illustration shows how RTEC performs CE recognition. To support real-time reasoning, at each query time $Q_i$ all SDEs that took place before or at $Q_i{-}\WM$ are discarded. To handle efficiently delayed SDEs and SDE revision, CE intervals within $\WM$ are computed from scratch. At $Q_i$, the computed maximal CE intervals are those that can be derived from SDEs that occurred in the interval $(Q_{i}{-}\WM, Q_{i}]$, as recorded at time $Q_i$. For completeness, RTEC amalgamates the computed intervals to any intervals ending at $Q_i{-}\WM$. More details about CE recognition in RTEC may be found at \cite{artikisDEBS12}.

\begin{figure*}[t]
        \centering
        \begin{subfigure}[b]{.51\textwidth}
                \centering
                \includegraphics[width=\textwidth]{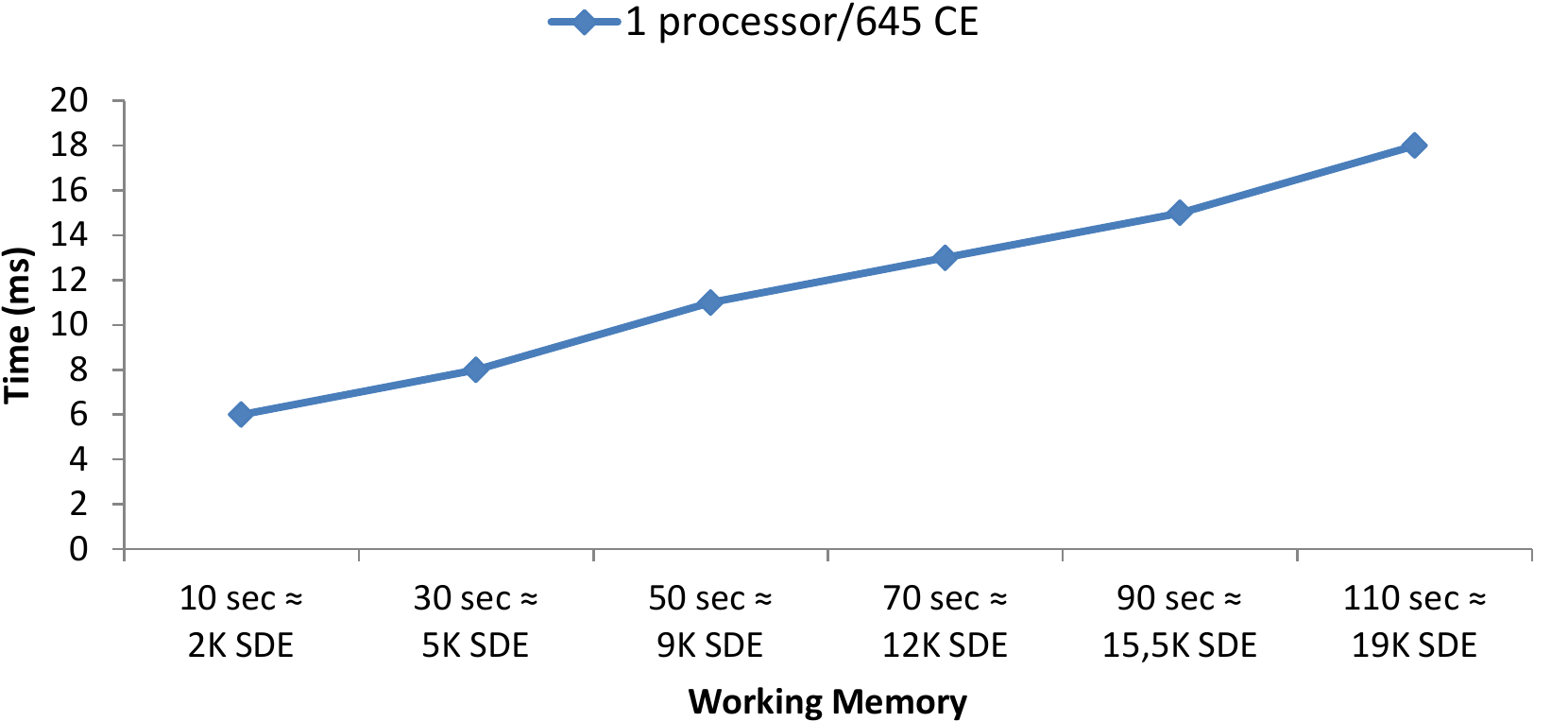}
                \caption{CE recognition for all 10 CAVIAR tracked entities.}
        \end{subfigure}%
        \begin{subfigure}[b]{.5\textwidth}
                \centering
                \includegraphics[width=\textwidth]{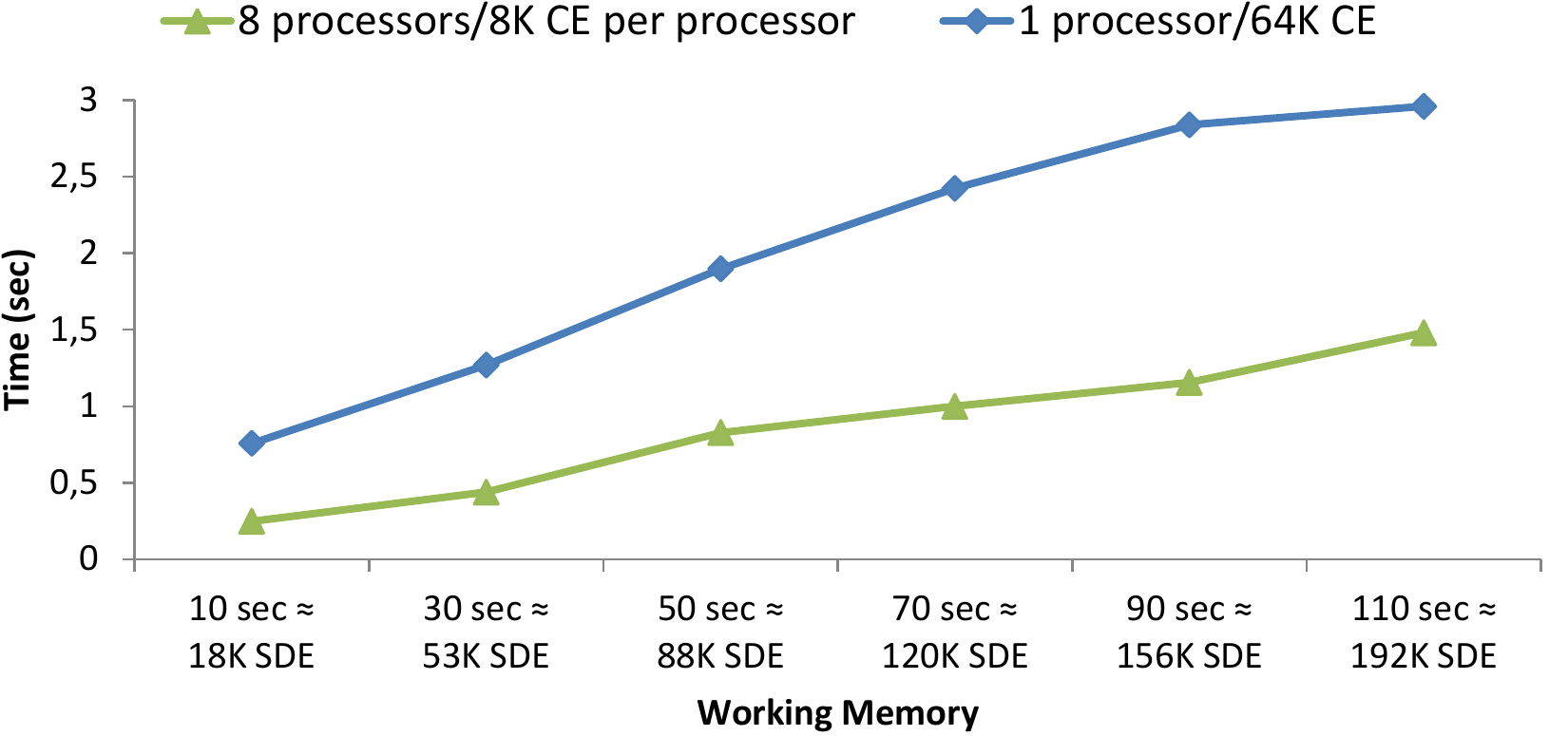}
                \caption{CE recognition for 100 tracked entities.}
        \end{subfigure}
\caption{Event Recognition for Public Space Surveillance.\vspace{10mm}}\label{fig:PSS-large-scale-experiments}
\end{figure*}

\section{Experimental Results}\label{sec:experiments}

We present experimental results on the public space surveillance application. The experiments were performed on a computer with eight Intel i7 950@3.07GHz processors and 12GiB RAM, running Ubuntu Linux 12.04 and YAP Prolog 6.2.2. 
Each CE recognition time displayed in this section is the average of 30 runs.
%
%
We use the CAVIAR benchmark dataset consisting of 28 surveillance videos of a public space \url{<http://groups.inf.ed.ac.uk/vision/CAVIAR/CAVIARDATA1>}. The videos are staged---actors walk around, sit down, meet one another, leave objects behind, etc. Each video has been manually annotated by the CAVIAR team in order to provide the ground truth for `short-term activities', i.e.~activities taking place in a short period of time detected on individual video frames. (The frame rate in CAVIAR is 40 ms.) The short-term activities of CAVIAR concern an entity (person or object) entering or exiting the surveillance area, walking, running, moving abruptly, being active or inactive. The CAVIAR team has also annotated the 28 videos with `long-term activities': a person leaving an object unattended, two people meeting, moving together and fighting. 
Short-term activities can be viewed as SDEs while long-term activities can be viewed as CEs. Consequently, the input to RTEC in this case study includes the set of annotated short-term activities, and the output is a set of recognised long-term activities. 
The CE definitions and the datasets on which the experiments were performed are available with the RTEC code.

\textbf{CE recognition for multiple pairs of entities. }
Figure \ref{fig:PSS-large-scale-experiments}(a) shows the results of experiments concerning all 45 pairs of the 10 entities tracked in the CAVIAR dataset. (In CAVIAR each CE concerns a pair of entities.) On average, 179 SDEs are detected per sec. We used a single processor for CE recognition concerning all 45 tracked pairs. That requires computing and storing the intervals of 645 CEs. 
%
%
We varied $\WM$ from 10~sec ($\approx$2,000 SDEs) to 110~sec ($\approx$19,000~SDEs). 
The inter-query step is set to 5 sec ($\approx$1,000 SDEs). 
In all settings shown in Figure \ref{fig:PSS-large-scale-experiments}(a), RTEC performs real-time CE recognition.

\textbf{Larger datasets. }
We constructed a larger dataset by taking ten copies of the original CAVIAR dataset with new identifiers for the tracked entities in each copy. The resulting dataset has 100 tracked entities, i.e.~4,950 entity pairs, while on average 1,800 SDEs take place per sec. According to the use case survey of the Event Processing Technical Society \cite{bizzaro11}, in the resulting dataset there are more SDEs per sec than in most applications. 
First, we used a single processor for CE recognition. In this case, the intervals of approximately 64,000 CEs were computed and stored. Second, we used all eight processors of the computer in parallel. Consequently, each instance of RTEC running on a processor computed and stored the intervals of approximately 8,000 CEs.
We emphasize that the input data was the same in all sets of experiments: each processor receives SDEs coming from \emph{all} tracked entities---i.e.~there was no SDE filtering to restrict the input relevant for each processor. We rely only on the indexing mechanism of RTEC to pick out relevant SDEs from the stream.
RTEC employs a very simple indexing mechanism: it merely exploits YAP Prolog's standard indexing on the functor of the first argument of the head of a clause.

As in the previous set of experiments, the inter-query step is set to 5 sec, while the size of the $\WM$ varies from 10 to 110~sec. In this case, however, step includes approximately 9,000 SDEs, and \WM\ varies from 18,000  to 192,000 SDEs. 
Figure \ref{fig:PSS-large-scale-experiments}(b) shows the average CE recognition times. 
In all cases RTEC performs real-time CE recognition. 
Figure \ref{fig:PSS-large-scale-experiments}(b) also shows that we can achieve significant performance gain by running RTEC in parallel on different processors.  Such a gain is achieved without requiring SDE filtering.
%

\section{Discussion}\label{sec:summary}

We presented RTEC, an Event Calculus dialect with novel implementation techniques that allow for efficient CE recognition, scalable to large numbers of SDEs and CEs. 
RTEC remains efficient and scalable in applications where SDEs arrive with a (variable) delay from, or are revised by, the SDE detection systems: it can update the already recognised CEs, and recognise new CEs, when SDEs are arrive with a delay or following revision.

RTEC has a formal, declarative semantics as opposed to most complex event processing languages, several data stream processing and event query languages, and most commercial production rule systems. Furthermore, RTEC has available the power of logic programming and thus supports atemporal reasoning and reasoning over background knowledge (as opposed to e.g.~\cite{arasu06, dousson07, kramer09, cugola10}), has built-in axioms for complex temporal phenomena (as opposed to \cite{davis07CVPR,anicic12}), explicitly represents CE intervals and thus avoids the related logical problems (as opposed to e.g.~\cite{mahbub11, dousson07, cugola10, kakas05}), and supports out-of-order SDE streams (as opposed to \cite{gyllstrom07, ding08, cugola10, dindar11, li11, paschke08}). 
Concerning the Event Calculus literature, a key feature of RTEC is that it includes a windowing technique. In contrast, no Event Calculus system (including e.g.~\cite{chittaro96,chesani10,montali14,paschke08,cervesato00}) `forgets' or represents concisely the SDE history. 

The `Cached Event Calculus' \cite{chittaro96} performs \emph{update-time} reasoning: it computes and stores the consequences of a SDE as soon as it arrives. Query processing, therefore, amounts to retrieving the appropriate CE intervals from the computer memory.
When a maximal interval of a fluent is asserted or retracted due to a delayed SDE, the assertion/retraction is propagated to the fluents whose validity may rely on such an interval. 
E.g.~$\mathit{propagateAssert([T_1,T_2], U)}$ in the Cached Event Calculus checks whether there are new initiations as a result of asserting the interval $(T_1, T_2]$ of fluent $\mathit{U}$.
In particular, $\mathit{propagateAssert}$ checks whether: (1) the asserted fluent $\mathit{U}$ is a condition for the initiation of a fluent $F$ at the occurrence of event $E$, (2) the occurrence time $T$ of $E$ belongs to $(T_1, T_2]$, and (3) there is not already a maximal interval for $F$ with $T$ as its starting point. If the above conditions are satisfied, $\mathit{propagateAssert}$ recursively calls $\mathit{updateInit(E,T,F)}$ in order to determine if $F$ is now initiated at $T$, and if it is, to update the fluent interval database accordingly. 

$\mathit{propagateAssert}$ also checks whether there are new terminations as a result of a fluent interval assertion, while $\mathit{propagateRetract}$ checks whether there are new initiations and terminations as a result of a fluent interval retraction.
The cost of $\mathit{propagateAssert}$ and $\mathit{propagateRetract}$ is very high, especially in applications where the CE definitions include many rules with several fluents that depend on several other fluents. Furthermore, this type of reasoning is performed very frequently.
RTEC avoids the costly checks every time a fluent interval is asserted/retracted due to delayed SDE arrival/revision. We found that in RTEC it is more efficient, and simpler, to discard at each query time $Q_i$, all intervals of fluents representing CEs in $\mathit{(Q_i{-}WM, Q_i]}$ and compute from scratch all such intervals given the SDEs available at $Q_i$ and detected in $\mathit{(Q_i{-}WM, Q_i]}$. 


For further work, we are developing techniques, based on abductive-inductive logic programming, for automated generation and refinement of CE definitions from very large datasets, with the aim of minimising the time-consuming and error-prone process of manual CE definition construction \cite{DBLP:journals/corr/KatzourisAP14}.
We are also porting RTEC into probabilistic logic programming frameworks, in order to deal with various types of uncertainty, such as imperfect CE definitions, incomplete and erroneous SDE streams \cite{kimmig11}. 

\ack We would like to thank the anonymous reviewers for their very helpful comments. This work has been partly funded by the EU FP7 project SPEEDD (619435).

\end{document}